\documentclass{article}
\usepackage[utf8]{inputenc}
\usepackage{authblk}
\usepackage{setspace}
\usepackage[margin=1.25in]{geometry}
\usepackage{graphicx}
\graphicspath{ {./figures/} }
\usepackage{subcaption}
\usepackage{amsmath}
\usepackage{amssymb}
\usepackage{amsfonts}
\usepackage{tipa}
\usepackage{hyperref}

\usepackage[style=nejm, 
citestyle=numeric-comp,
sorting=none]{biblatex}
\title{QueEn: A Large Language Model for Quechua-English Translation}

\newcommand{\equalcontrib}{\textsuperscript{*}}

\author[1]{Junhao Chen\thanks{Co-first author}}
\author[1]{Peng Shu\equalcontrib}
\author[1]{Yiwei Li}
\author[1]{Huaqin Zhao}
\author[1]{Hanqi Jiang}
\author[1]{Yi Pan}
\author[1]{Yifan Zhou}
\author[1]{Zhengliang Liu}
\author[2]{Lewis C Howe}
\author[1]{Tianming Liu\thanks{Corresponding author}}

\affil[1]{School of Computing, The University of Georgia, Athens 30602, USA}
\affil[2]{Department of Linguistics, The University of Georgia, Athens 30602, USA}

\date{}

\begin{document}

\maketitle
\begin{abstract}
Recent studies show that large language models (LLMs) are powerful tools for working with natural language, bringing advances in many areas of computational linguistics. However, these models face challenges when applied to low-resource languages due to limited training data and difficulty in understanding cultural nuances. In this paper, we propose QueEn, a novel approach for Quechua-English translation that combines Retrieval-Augmented Generation (RAG) with parameter-efficient fine-tuning techniques. Our method leverages external linguistic resources through RAG and uses Low-Rank Adaptation (LoRA) for efficient model adaptation. Experimental results show that our approach substantially exceeds baseline models, with a BLEU score of 17.6 compared to 1.5 for standard GPT models. The integration of RAG with fine-tuning allows our system to address the challenges of low-resource language translation while maintaining computational efficiency. This work contributes to the broader goal of preserving endangered languages through advanced language technologies.
\end{abstract}

\section{Introduction}
Large Language Models (LLMs) have achieved remarkable success across a wide range of tasks \cite{jiang2024oraclesage,xu2024towards,shu2024transcending,wang2024legal,kim2024echofm,wu2024robot}, revolutionizing natural language understanding, generation, and reasoning. These models, powered by advanced architectures and massive datasets, have become the backbone of applications in fields such as content creation, code generation, and conversational AI\cite{chen20243d,shu2024llms,dam2024complete}. In parallel, multimodal LLMs, which integrate vision and language, are rapidly advancing, enabling capabilities such as image captioning, visual question answering, and multimodal reasoning\cite{bianco2023improving,guo2023images,pan2024eg}. Despite these significant achievements, the application of LLMs in low-resource regions, particularly for low-resource languages, remains underexplored and underdeveloped.
One of the key challenges facing LLMs in low-resource language translation is the lack of sufficient fine-tuning data\cite{hasan2024large}. Many low-resource languages suffer from limited publicly available corpora, making it difficult to train or fine-tune models effectively. Furthermore, LLMs often demonstrate poor performance in zero-shot settings for these languages\cite{shu2024transcending}, as their training predominantly relies on high-resource languages, leading to a significant disparity in linguistic representation. This results in inadequate translations, reduced contextual understanding, and a lack of cultural nuance in the generated outputs.
Recognizing these limitations, there is a growing emphasis on improving the performance and application of LLMs in low-resource languages\cite{lankford2023adaptmllm,cahyawijaya2024llms}. Researchers are increasingly focusing on developing novel methods to address the scarcity of data, such as leveraging synthetic data generation, cross-lingual transfer learning, and community-driven data collection efforts. By overcoming these challenges, the goal is to enable LLMs to bridge the gap in linguistic equity and expand their utility to underserved regions, fostering greater inclusivity in AI advancements.

An endangered language is one that is at risk of falling out of use, typically because its speakers are shifting to using another language. Although the death of some language is inevitable, the number of endangered languages has actually rapidly increased. Ethnologue reported 7,099 languages in 2017, compared to the estimated 6,000 in 1992 by Krauss, who warned that 90\% of these languages could disappear by 2100 if current trends persist \cite{endangeredlangauge}. Quechua, one of the most widely spoken indigenous language families in the Americas, is nonetheless classified as vulnerable by UNESCO. Despite its historical significance as the language of the Inca Empire, its current condition varies widely across regions. In countries like Peru, Bolivia, and Ecuador, Quechua retains a considerable number of speakers, but the language faces significant challenges, including intergenerational transmission gaps and diminishing domains of use. In urban areas, younger generations are increasingly adopting Spanish due to its perceived socio-economic advantages, leading to a decline in the number of fluent speakers. While national and regional governments in South America have taken steps to recognize and promote Quechua, such as integrating it into educational curricula and granting it official status, these efforts have been inconsistent and often underfunded. As a result, the long-term viability of Quechua remains uncertain without more robust and sustained revitalization measures.

Translating Quechua presents significant challenges due to its status as a low-resource language. The scarcity of digital resources, such as comprehensive corpora and linguistic tools, hampers the development of effective machine translation systems. The complex morphology of Quechua, characterized by agglutination and polysynthetic structures, further complicates translation efforts. Words often consist of multiple morphemes, each carrying distinct meanings, leading to a high degree of inflection and variability. This complexity poses difficulties for neural machine translation models, which struggle to accurately process and generate Quechua text. Additionally, the limited availability of parallel Quechua-Spanish corpora restricts the training of robust translation models, resulting in lower performance compared to high-resource language pairs. Efforts to address these challenges include morphological segmentation techniques and the development of specialized translation engines, but significant obstacles remain to achieve high-quality Quechua translation.

In this paper, we propose a Retrieval-Augmented Generation (RAG)-based fine-tuning method to address the translation challenges of Quechua. Our approach leverages the strengths of retrieval-augmented techniques to enhance the fine-tuning dataset by incorporating relevant, high-quality data retrieved from external resources. This enriched training process significantly improves the translation performance compared to zero-shot methods. Our experimental results demonstrate that the proposed method outperforms baseline models such as GPT-4o and LLaMA 405B, establishing its effectiveness in low-resource language translation. These findings highlight the potential of RAG-based fine-tuning to bridge the performance gap for underrepresented languages like Quechua.

\section{Related Works}
Research on low-resource languages (LRLs) has gained increasing attention in recent years, particularly due to the challenges these languages present in natural language processing (NLP) tasks. Unlike high-resource languages such as English or Mandarin, which benefit from large datasets and extensive linguistic resources, LRLs often suffer from a lack of annotated corpora, lexicons, and linguistic tools. To overcome these challenges, the field of machine translation (MT) has evolved significantly over the decades, progressing through several distinct paradigms before the advent of LLMs. These paradigms include rule-based, statistical, and neural machine translation, each contributing foundational methods and insights that shaped the current state of translation technology.

\subsection{Rule-Based Machine Translation}
Rule-Based Machine Translation (RBMT) systems\cite{shiwen2014rule} were the earliest attempts at automated translation, emerging in the 1950s and 1960s during the formative years of computational linguistics. These systems relied heavily on linguistic theories and human-crafted rules to model the grammar and syntax of the source and target languages. The central idea was to use bilingual dictionaries for word-level mappings and predefined rules to handle grammatical transformations. Early RBMT systems, such as the Georgetown-IBM experiment (1954)\cite{hutchins2004georgetown}, demonstrated the feasibility of MT by translating a set of Russian sentences into English, albeit under highly constrained conditions.
RBMT methods followed one of three primary strategies: direct\cite{shiwen2014rule}, transfer-based\cite{llitjos2005framework}, or interlingua-based translation\cite{richens1958interlingual}. Direct translation systems performed word-by-word substitution and were typically limited to specific language pairs with similar syntactic structures. Transfer-based systems introduced an intermediary step, where the source language was first converted into an abstract syntactic representation before being mapped to the target language. Interlingua-based systems took this step further, converting the source language into a universal, language-agnostic representation, enabling translation to multiple target languages from the same source structure.

\subsection{Statistical Machine Translation}
The emergence of Statistical Machine Translation (SMT)\cite{lopez2008statistical} in the late 1980s and early 1990s marked a paradigm shift in MT research. Unlike RBMT, SMT systems relied on data-driven approaches, using large bilingual corpora to learn translation patterns automatically. The foundation of SMT was laid by researchers at IBM with the introduction of probabilistic models\cite{brown1993mathematics}, such as the IBM Models 1-5\cite{collins2011statistical}, which focused on word alignment and translation probability estimation. These models demonstrated that translation could be framed as a statistical optimization problem, where the goal was to maximize the probability of a target sentence given a source sentence.
Phrase-based SMT\cite{okuma2008introducing}, an extension of word-based models, became the dominant approach in the early 2000s. By treating phrases (sequences of words) as the basic units of translation, phrase-based SMT systems captured more contextual information and achieved greater fluency compared to word-level models. Additionally, SMT systems integrated several components, including language models, translation models, and reordering models, to optimize the overall quality of translations.
The strengths of SMT lie in its adaptability and ability to learn directly from data, reducing the reliance on handcrafted rules. However, its performance was heavily dependent on the availability of large, high-quality parallel corpora. For low-resource languages, this posed a significant challenge, as the lack of sufficient training data often led to poor translations. SMT systems also struggled with long-distance dependencies and syntactic ambiguities, as their models were primarily designed for local phrase-level predictions.
To mitigate some of these limitations, techniques such as domain adaptation, unsupervised alignment, and hierarchical phrase-based models\cite{chiang2005hierarchical} were developed. While these methods improved SMT's robustness, they still fell short of delivering human-level translation quality, particularly for morphologically complex and underrepresented languages.
While RBMT systems offered early successes, they were highly labor-intensive to develop, requiring detailed linguistic knowledge for each new language pair. Their reliance on rigid, deterministic rules often produced translations that lacked fluency and naturalness, particularly for idiomatic expressions or complex sentence structures. Moreover, these systems struggled with scalability as the number of languages and linguistic variations increased, leading to a decline in their popularity as statistical approaches gained traction.

\subsection{Neural Machine Translation}
The advent of deep learning in the 2010s revolutionized MT with the introduction of Neural Machine Translation (NMT)\cite{stahlberg2020neural}. Unlike SMT, which relied on multiple components for different aspects of translation, NMT adopted an end-to-end approach, enabling models to learn the entire translation process from input to output. Early NMT systems were based on the encoder-decoder architecture using recurrent neural networks (RNNs)\cite{zhang2017context}. The encoder transformed the input sentence into a fixed-length vector representation, which the decoder then used to generate the target sentence. This method demonstrated significant improvements in translation fluency and contextual understanding.
The introduction of the attention mechanism by Bahdanau\cite{bahdanau2014neural} addressed one of the key limitations of traditional encoder-decoder models: their inability to handle long sentences effectively. Attention mechanisms allowed the decoder to focus on relevant parts of the input sequence at each step of the translation, resulting in more accurate and coherent outputs. This innovation laid the groundwork for further advancements in NMT.
The development of the Transformer architecture by Vaswani\cite{vaswani2017attention} marked another breakthrough in NMT. By replacing RNNs with self-attention mechanisms, Transformers enabled parallel processing of input sequences, significantly improving both translation quality and computational efficiency. Transformer-based models, such as OpenNMT\cite{klein2017opennmt}, Marian\cite{junczys2018marian}, and the NMT systems used in Google Translate\cite{wu2016google}, became the new standard for MT, achieving near-human-level performance for many high-resource language pairs.
Despite their success, NMT systems faced challenges similar to SMT in low-resource settings. The effectiveness of NMT depended heavily on large-scale parallel corpora, and its performance degraded significantly for languages with limited training data. To address these challenges, researchers explored techniques such as transfer learning, multilingual NMT, and back-translation. Transfer learning leveraged pre-trained models on high-resource languages to improve performance on low-resource languages, while multilingual NMT trained models to handle multiple languages simultaneously, enabling knowledge transfer across language pairs. Back-translation involved generating synthetic parallel data by translating target language text into the source language, effectively augmenting the training dataset.
While these methods partially alleviated the challenges of low-resource MT, they were not sufficient to close the performance gap, particularly for underrepresented languages with unique linguistic characteristics and limited digital resources. This persistent issue underscored the need for more innovative approaches, setting the stage for the application of LLMs and retrieval-augmented techniques in low-resource language translation.

\subsection{Quechua}
Quechua, often referred to as Runa Simi (the "language of the people"), is one of the most significant indigenous languages of the Americas\cite{hornberger2004quechua,hornberger2001reversing}. Once the lingua franca of the mighty Inca Empire, Quechua has endured centuries of colonial disruption and marginalization, yet it remains a vital thread in the cultural and linguistic fabric of South America. Today, millions of speakers across several Andean nations continue to use Quechua in daily life, preserving its rich heritage and ensuring its survival into the 21st century.

The linguistic structure of Quechua provides a fascinating glimpse into its complexity and adaptability. Its phonological, morphological, syntactic, semantic, and pragmatic features\cite{adelaar2020morphology,faller2002semantics} reveal the language’s uniqueness and its rich potential for nuanced expression.
Quechua's phonological system is relatively simple, featuring three vowel sounds (*/a/, */i/, /u/) and a modest consonant inventory. In some dialects, vowels may shift to harmonize with neighboring sounds, a phenomenon known as vowel harmony.
Quechua’s consonant inventory includes stops, fricatives, nasals, and laterals. A key feature of the language is the use of aspirated and glottalized stops, especially in Southern Quechua dialects, distinguishing it from other Andean languages. The phonemic inventory includes:
\begin{itemize}
    \item Stops: /p/, /t/, /k/, and their aspirated and glottalized counterparts.
    \item Fricatives: /s/, /h/.
    \item Nasals: /m/, /n/, /\textipa{N}/ (palatal nasal).
    \item Laterals: /l/.
\end{itemize}
Quechua syllables generally follow a CV (consonant-vowel) pattern, making the language syllable-timed and rhythmic in its spoken form. Clusters of consonants are rare, which simplifies pronunciation.
Quechua is an agglutinative language, meaning it constructs words by stringing together morphemes, each carrying a specific grammatical or semantic function. This highly systematic morphology is one of the defining characteristics of the language.Most Quechua words consist of a root followed by one or more suffixes.Suffixes can convey a wide array of meanings, including:
\begin{itemize}
    \item Pluralization: -kuna (plural)
    \item Tense: -rqa (past), -sha (progressive)
    \item Possession: -yki (your), -n (his/her/its)
\end{itemize}
Quechua follows a Subject-Object-Verb (SOV) word order, although this can vary slightly for emphasis or poetic expression.
Semantics in Quechua is deeply tied to its cultural and environmental context. The language offers sophisticated ways of expressing relationships, time, and evidentiality.

Quechua holds a paradoxical position in modern society. While it is recognized as an official language in countries like Peru, Bolivia, and Ecuador, its speakers often face social and economic marginalization. Urbanization and the dominance of Spanish have led to language shift, with younger generations increasingly favoring Spanish over Quechua.

\subsection{Large Language Models}
The advent of large language models (LLMs), such as GPT-3\cite{brown2020language}, GPT-4\cite{achiam2023gpt}, and LLaMA\cite{touvron2023llama}, has revolutionized natural language processing (NLP) by enabling context-aware and coherent text generation across diverse tasks. These models, trained on massive corpora encompassing multiple languages, have shown remarkable potential in addressing the challenges of low-resource language translation. Unlike traditional machine translation approaches that rely heavily on parallel corpora, LLMs can leverage in-context learning, zero-shot, and few-shot capabilities to produce translations with minimal training data. This makes them particularly well-suited for languages with limited resources, such as Quechua, where large-scale parallel corpora are often unavailable.

Building on these advances, innovative methods like retrieval-augmented generation (RAG)\cite{lewis2020retrieval} and prompt engineering\cite{wang2023review} have emerged as powerful tools to enhance LLM performance for low-resource language translation. By integrating linguistic resources directly into the translation pipeline, these methods enable LLMs to effectively utilize dictionaries, grammar guides, and other linguistic tools to produce accurate and contextually appropriate translations. In the context of Quechua-English translation, such techniques are poised to address longstanding challenges by combining the generative power of LLMs with retrieval-based mechanisms and linguistic insights.

In fine-tuning low-resource language translation systems for the Quechua language, similar data augmentation techniques as those employed for the Manchu-Korean translation in the Mergen model can be applied\cite{seo2023mergen}. In the context of Quechua-English translation, the QueEn project developed a Quechua-English parallel dataset to aid machine translation research between these two languages\cite{ortega2020neural}. However, given the extremely low-resource nature of Quechua, with only 14,000 sentence pairs available, leveraging data augmentation through synonym generation and back-translation, as done in the Manchu-Korean project, could significantly enhance performance. By employing methods such as training multiple versions of GloVe embeddings\cite{tifrea2018poincar} or integrating multilingual pre-trained models like BERT\cite{kenton2019bert}, researchers can expand the dataset size and improve translation quality. Data augmentation, combined with transfer learning and semi-supervised methods, could similarly boost the performance of Quechua-English translation systems, as shown by the use of back-translation and monolingual data to improve machine translation models in the QueEn project.
Incorporating data augmentation techniques into low-resource language processing has shown significant promise in enhancing machine translation performance, especially for endangered languages. A notable example is the Manchu-Korean machine translation system, Mergen, which faced a scarcity of parallel data. To overcome this challenge, the researchers employed a data augmentation strategy using GloVe embeddings to expand the parallel dataset. By generating synonym replacements through word embeddings, they were able to augment their training data significantly. This method resulted in a remarkable improvement in translation quality, with a significant BLEU score increase.Data augmentation, especially through synonym generation and word replacement techniques, can thus address the inherent data scarcity in low-resource language contexts, improving translation and NLP outcomes by increasing the size and variability of the training corpus. This strategy is particularly useful when working with languages that have limited digitized text or parallel corpora.
In contrast to traditional fine-tuning approaches for low-resource language translation, prompt engineering offers an innovative and training-free alternative, as demonstrated in the LINGOLLM\cite{zhang2024hire} approach. This method involves using large language models (LLMs) like GPT-4 by incorporating in-context linguistic descriptions, such as dictionaries and grammar books, directly into the prompt. For instance, LINGOLLM uses morphological analyzers to preprocess the input by breaking it down into morphemes, which are then glossed using a dictionary. The model is further guided by grammar rules to perform accurate translations without requiring extensive training data. In a low-resource language context, such as translating Quechua, this method could be particularly advantageous, especially when rich linguistic resources like grammar books and dictionaries are available but large parallel corpora are not. This prompt-based approach achieves significant improvements in translation tasks across several endangered languages, elevating GPT-4’s BLEU score from near-zero to over 10.5 on average. The ability to harness linguistic resources efficiently using prompt engineering presents a flexible and scalable solution for low-resource language translation.

\section{Methods}
\begin{figure}
  \centering
  \includegraphics[width=1.0\linewidth]{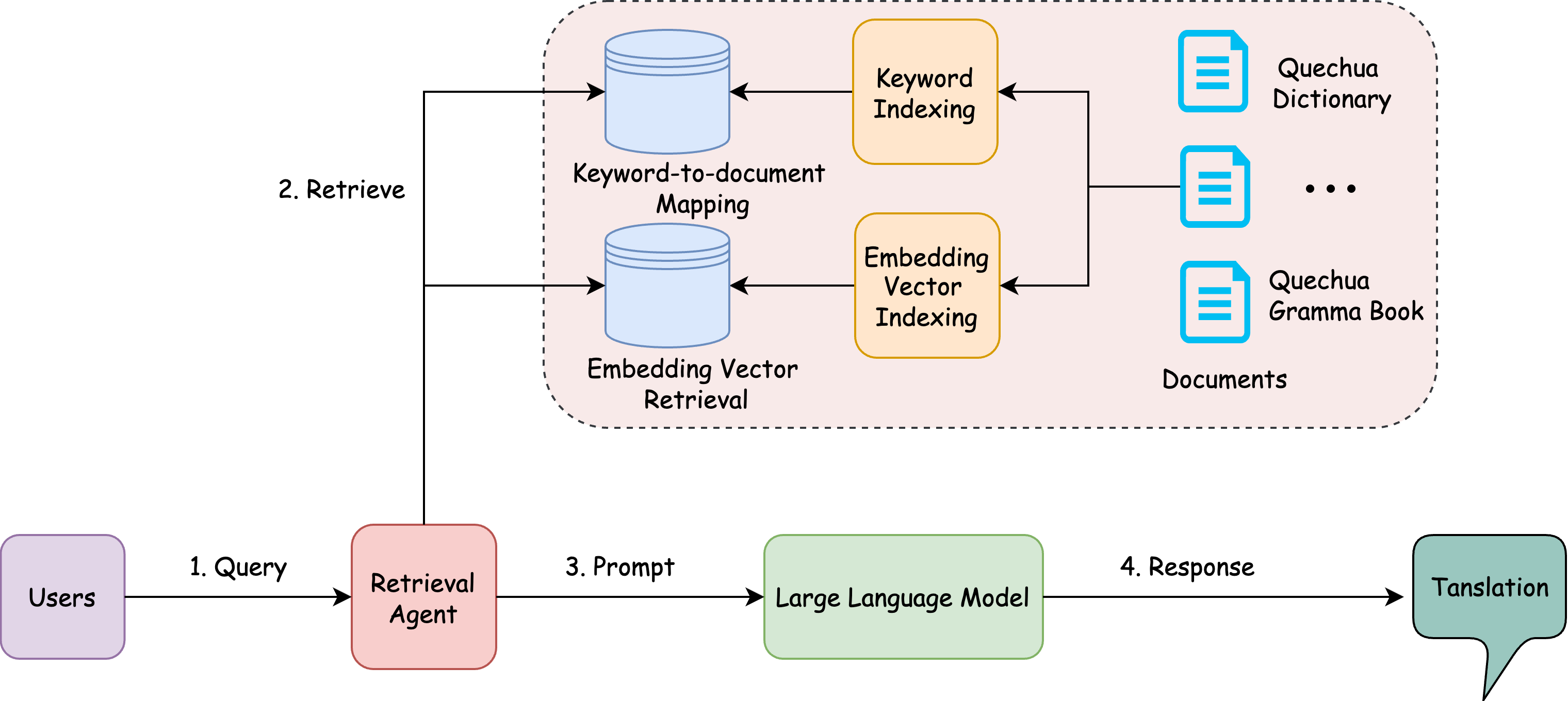}
  \caption{Illustrating a retrieval-augmented generation (RAG) architecture: Documents are indexed using both keyword and embedding vector methods, stored in separate databases. A retrieval agent accesses these indexes to provide relevant information, which is then processed by a GPT-4 model to deliver responses to users.}
  \label{fig:rag}
\end{figure}

In this work, we introduce a robust framework, shown in Figure \ref{fig:rag}, for translating Quechua into English using a large language model (LLM) enhanced with fine-tuning, retrieval-augmented generation (RAG), and data augmentation techniques. Fine-tuning is used to adapt the pre-trained LLM to the specific characteristics of Quechua, a low-resource language, while minimizing resource constraints through parameter-efficient strategies. Retrieval-augmented generation further bolsters translation quality by incorporating external linguistic resources into the generation process. Additionally, data augmentation is applied to expand the dataset and address data scarcity challenges, thereby improving model robustness and generalization. Each of these components is elaborated upon in the subsequent subsections.
\subsection{Fine-tuning}
Fine-tuning is a key method for adapting pre-trained LLMs to specific tasks or domains. After an LLM has been pre-trained on massive, general-purpose datasets, fine-tuning allows the model to refine its knowledge using a smaller, more focused dataset. This process customizes the model to excel in specific applications, such as translation, sentiment analysis, or domain-specific text generation.
Fine-tuning typically involves continuing the training process of the pre-trained model using task-specific data. The objective is to adjust the model's weights in a way that optimizes performance for the target task while preserving its general knowledge. Fine-tuning methods can be broadly classified into full fine-tuning\cite{lv2023full} and parameter-efficient fine-tuning\cite{ding2023parameter}:
\begin{itemize}
  \item  Full Fine-Tuning: All the parameters of the model are updated during fine-tuning. While effective, this approach is computationally expensive and requires significant resources, particularly for large-scale models.
  \item Parameter-Efficient Fine-Tuning: Only a subset of the model's parameters is updated, reducing computational costs. Techniques such as Low-Rank Adaptation (LoRA)\cite{hu2021lora} and adapter layers are examples of this category.
\end{itemize}

Among these methods, parameter-efficient fine-tuning has gained prominence for its ability to adapt LLMs with minimal resource usage. In this context, Low-Rank Adaptation (LoRA)\cite{hu2021lora} has emerged as a highly effective approach. LoRA is a parameter-efficient fine-tuning technique designed to minimize the computational and memory overhead of adapting LLMs. Instead of updating all model parameters, LoRA introduces a small number of trainable parameters that are added to the existing model structure.
LoRA assumes that the weight updates required for fine-tuning can be approximated by a low-rank decomposition. Let $ W \in \mathbb{R}^{d \times k} $ represent a weight matrix in the pre-trained model. During fine-tuning, instead of directly updating $W$, LoRA approximates the weight update

\begin{equation}
\Delta W = A \cdot B
\end{equation}

where $A \in \mathbb{R}^{d \times r}$ and $ B \in \mathbb{R}^{r \times k}$ are low-rank matrices, with $r \ll \min(d, k)$. The rank $r$ is a hyperparameter that controls the trade-off between computational efficiency and fine-tuning capacity. The updated weight matrix becomes:

\begin{equation}
W' = W + \Delta W = W + A \cdot B
\end{equation}

This decomposition significantly reduces the number of trainable parameters, as the number of parameters in $A$ and $B$ is much smaller than the number of parameters in $w$. By only training the low-rank matrices $A$ and $B$, LoRA dramatically reduces the memory and computational requirements of fine-tuning. The learned $A$ and $B$ matrices can be stored and reused for different tasks, making LoRA particularly useful in multi-task or multi-domain scenarios. Since the base weights $W$ remain fixed, LoRA ensures that the general knowledge of the pre-trained model is retained.
In this paper LoRA is applied to specific weight matrices within the Transformer architecture, such as the query and value projection matrices in the self-attention mechanism. For a given weight matrix $W_q$ in the query projection, the updated projection becomes:

\begin{equation}
W_q' = W_q + A_q \cdot B_q
\end{equation}

Here, $A_q$ and $B_q$ are low-rank matrices specific to the query projection. Similar updates can be applied to other matrices, such as $w_v$ for the value projection.
\subsection{Retrieval-augmented Generation(RAG)}
We leverage Retrieval-Augmented Generation (RAG) to enhance the translation of Quechua into English. Figure \ref{fig:rag} demonstrates the complete framework. The workflow begins by converting key resource documents, such as a Quechua-to-English dictionary and Quechua grammar guide, into structured databases. These databases are optimized for efficient information retrieval through dual mechanisms: keyword-based indexing and embedding-based vector indexing. During translation, the system retrieves relevant linguistic and grammatical information from these databases to augment the LLM with contextually relevant data, enabling the generation of accurate translations. Specifically, for a user query \( q \), the system retrieves supporting information \( D_q \), which is then used to construct a prompt \( P \) for the LLM. The model generates a response \( R \), synthesizing the retrieved context to improve translation quality. The process can be represented as:

\[
q \xrightarrow{\text{Retrieval}} D_q \xrightarrow{\text{Prompt Construction}} P \xrightarrow{\text{LLM}} R
\]

Here, \( q \) represents the input query, \( D_q \) is the set of retrieved documents or entries, \( P \) is the constructed prompt incorporating \( D_q \), and \( R \) is the final translation output.

Our approach integrates a dual retrieval mechanism to optimize the translation process. First, a \textbf{keyword-based index} allows efficient lookups by identifying exact matches between query terms and dictionary entries. Let \( \text{KeywordIndex}(D) \) represent the keyword-to-document mapping, where \( D \) is the set of all documents. For a query \( q \), the keyword retrieval process can be expressed as:

\[
D_q^{\text{keyword}} = \{ d \in D \mid q \cap \text{Keywords}(d) \neq \emptyset \}
\]

Second, when exact keyword matches are unavailable, an \textbf{embedding-based retrieval} mechanism is employed. Each document \( d \) and query \( q \) is encoded into a semantic vector representation using a function \( f_{\text{embed}}(\cdot) \), such that:

\[
v_d = f_{\text{embed}}(d), \quad v_q = f_{\text{embed}}(q)
\]

The top \( K \) most similar entries are retrieved based on cosine similarity:

\[
\text{Similarity}(v_q, v_d) = \frac{v_q \cdot v_d}{\Vert v_q \Vert \, \Vert v_d \Vert}
\]

The retrieved documents using vector-based retrieval are:

\[
D_q^{\text{vector}} = \{ d \in D \mid \text{Similarity}(v_q, v_d) \text{ is among the top } K \}
\]

The final retrieved set \( D_q \) is a combination of both methods:

\[
D_q = D_q^{\text{keyword}} \cup D_q^{\text{vector}}
\]

Finally, the retrieved set \( D_q \) is integrated into the LLM prompt \( P \), and the response \( R \) is generated:

\[
P = \text{ConstructPrompt}(q, D_q), \quad R = \text{LLM}(P)
\]

This hybrid retrieval mechanism ensures that both precise matches and semantically relevant content are considered, improving translation accuracy. By combining these retrieval techniques with the LLM’s generative capabilities, we address challenges specific to low-resource languages like Quechua, enabling accurate and contextually enriched translations.


\section{Experiments}
\subsection{Data}
To test the performance of our translation model, this work utilizes a subset of the Siminchik corpus\cite{cardenas2018siminchik}, provided as part of the IWSLT2023 Low-resource Speech Translation Track\cite{ortega2020neural}. The dataset, referred to as que\_spa\_clean, comprises 1 hour and 40 minutes of clean speech in Quechua paired with Spanish translations. It includes raw transcriptions for Quechua and Spanish and true-cased Spanish translations, processed using a sacremoses Truecaser model trained on the WMT13 EN-ES dataset. The dataset is divided into validation and test sets, ensuring robust model evaluation.

Since the dataset provides translations from Spanish to Quechua, an additional translation step from Spanish to English is required. e applied ChatGPT-4 for Spanish-to-English translation, leveraging its advanced natural language processing capabilities. ChatGPT-4 has demonstrated high accuracy and contextual understanding in various linguistic tasks, making it a reliable tool for this intermediary translation step. To enhance the data and improve evaluation reliability, we generated multiple English translations for each Spanish sentence. These additional references help account for linguistic variability and ensure a more comprehensive evaluation of the English-to-Quechua translation performance. This approach enables us to evaluate English-to-Quechua translation scenarios effectively while maintaining high fidelity in the intermediate Spanish-English translation, ensuring the robustness and validity of our experiments.

\subsection{Evaluation Metrics}

To assess the performance and capabilities of our model in low-resource machine translation, we conducted experiments comparing it with GPT-4o and LLaMA 3.1 405B models. The objective was to evaluate the translation quality of these models, particularly for Quechua translations from English source sentences, using a robust evaluation framework. The experiment involved randomly selecting text from test datasets, where sentences were input sequentially into each model to generate translations in the target low-resource language. A combination of automated and human evaluation metrics was employed to ensure a comprehensive analysis of translation quality in terms of both lexical accuracy and semantic fidelity.

We used \textbf{ROUGE} \cite{lin2005recall} (Recall-Oriented Understudy for Gisting Evaluation) and \textbf{BLEU} \cite{papineni2002bleu} (Bilingual Evaluation Understudy) as core evaluation metrics. ROUGE measures the similarity between machine-generated translations and reference translations by analyzing overlapping n-grams, word sequences, and word pairs. Although commonly used in summarization tasks, ROUGE also provides valuable insights into translation fluency and accuracy. BLEU calculates n-gram precision between the generated and reference translations, offering a quantitative measure of how faithfully the output matches the reference. These metrics are particularly effective for surface-level evaluation, such as determining the overlap of words and phrases.

To capture deeper semantic similarities, we incorporated \textbf{BERTScore} \cite{zhang2019bertscore}, which leverages contextual embeddings from pretrained models to evaluate the semantic alignment between generated and reference translations. Unlike ROUGE and BLEU, which focus on exact lexical matches, BERTScore assesses meaning and context, making it especially suitable for low-resource languages where direct lexical matches may be limited.

\subsection{Experiment Results}
Both LLaMA and GPT exhibit low performance in traditional machine translation metrics such as BLEU and ROUGE, as shown in Table \ref{tab:example_table}. LLaMA achieves the lowest BLEU score and ROUGE, while plain GPT performs slightly better. These results indicate a significant lack of overlap between the generated translations and reference texts in terms of both n-gram precision (BLEU) and sentence-level structure (ROUGE). The scores reflect the inherent challenges of translating low-resource languages like Quechua, where limited training data and the linguistic complexity of the language hinder the models' ability to produce accurate surface-level translations.

The integration of Retrieval-Augmented Generation leads to substantial improvements in translation quality for both models. LLaMA + RAG achieves a BLEU score of 0.154 and a ROUGE score of 0.167, while GPT + RAG outperforms all other models with BLEU and ROUGE scores of 0.235 and 0.278, respectively. These results demonstrate that the retrieval component enhances the models' ability to align their outputs with the reference texts, addressing issues of lexical and structural alignment. The retrieval mechanism allows the models to access additional contextual information, improving their word and phrase-level accuracy while also capturing syntactic nuances more effectively. Despite their relatively low BLEU and ROUGE scores, all models maintain high BERT scores, reflecting their strength in preserving semantic meaning. Both RAG-enhanced models achieve the highest BERT scores (0.963), demonstrating their ability to retain the core semantic essence of the input text. This suggests that while traditional metrics highlight challenges in surface-level accuracy, BERT scores reveal the models' capacity to produce translations that align meaningfully with the reference texts. Among all models, GPT + RAG stands out as the most effective, benefiting from the retrieval mechanism to achieve superior performance in both lexical and semantic evaluations.
\begin{table}[ht]
    \centering
    \begin{tabular}{c c c c c}
        \hline
        Model  & Rouge Score & BLEU Score & BERT Score & \\ 
        \hline
        LLama   & 0.052 & 0.038 & 0.931 &  \\ 
        GPT   & 0.110 & 0.096 & 0.938 &  \\ 
        LLama + finetune  & 0.094 & 0.064 & 0.940 &  \\ 
        \hline
        LLama + RAG (ours) & \textbf{0.167} & \textbf{0.154} & \textbf{0.963} &  \\
        GPT + RAG (ours) & \textbf{0.278} & \textbf{0.235} & \textbf{0.963} & \\
        \hline
    \end{tabular}
    \caption{Results of Low-Resource Translation Experiments}
    \label{tab:example_table}
\end{table}

\section{Discussion}
Our work on QueEn demonstrates the potential of combining RAG with parameter-efficient fine-tuning for low-resource language translation. While our results show significant improvements over baseline models, several promising directions remain for future research and development.

First, the current implementation could be enhanced through the incorporation of more diverse linguistic resources. Future work should explore the integration of audio data and speech recognition capabilities, as Quechua has a strong oral tradition. This multimodal approach could help capture nuances of pronunciation and intonation that are crucial for proper understanding and translation. Additionally, the development of specialized embedding models for Quechua could improve the accuracy of our retrieval system.

Second, future work could focus on enhancing the model's contextual understanding and handling of complex grammatical structures. Given Quechua's agglutinative nature and rich morphological system~\cite{adelaar2020morphology}, developing specialized attention mechanisms or embedding layers that can better capture these linguistic features would be valuable. This could involve creating morpheme-aware tokenization strategies or incorporating explicit grammatical rule embeddings into the model architecture. Such improvements would help the model better handle Quechua's unique linguistic characteristics and produce more accurate translations

Third, community involvement and feedback mechanisms could be integrated into the system's development cycle. Future iterations could incorporate a human-in-the-loop approach, where native Quechua speakers can provide direct feedback on translation quality and help refine the model's understanding of cultural context. This collaborative approach would not only improve translation accuracy but also ensure that the technology serves the needs of the Quechua-speaking community. Such community-driven development is essential for cultural preservation, as it ensures that technological solutions respect and incorporate indigenous knowledge systems and cultural practices.

Finally, the techniques developed for QueEn could be adapted for other low-resource languages, particularly those facing similar challenges of cultural and linguistic preservation. The combination of RAG and parameter-efficient fine-tuning provides a promising framework for addressing the challenges of limited data and computational resources that are common across many endangered languages. Our approach could be particularly valuable for languages with strong oral traditions or complex morphological systems similar to Quechua. Creating a generalized version of our methodology could accelerate the development of translation systems for other indigenous languages, while respecting their unique cultural contexts and linguistic features. This generalization could play a crucial role in global efforts to preserve not just languages themselves, but the cultural heritage and traditional knowledge systems they embody.

\section{Conclusion}
In this paper, we introduced QueEn, a novel approach for Quechua-English translation that combines retrieval-augmented generation with parameter-efficient fine-tuning. Our experimental results demonstrate that this combination substantially improves translation quality compared to baseline models.

The implications of this work extend far beyond technical achievements. Effective translation tools for indigenous languages like Quechua can help bridge significant socioeconomic gaps in multilingual societies. By enabling better communication between Quechua-speaking communities and the wider world, such technologies can improve access to education, healthcare, and economic opportunities. This is particularly crucial in regions where language barriers have historically contributed to social and economic marginalization. Furthermore, our work contributes to the preservation of indigenous knowledge systems and cultural heritage, which are invaluable not only for the communities themselves but for humanity's collective understanding of diverse ways of knowing and being. As machine learning continues to advance, approaches like QueEn demonstrate how modern language technologies can promote linguistic diversity while fostering more inclusive and equitable societies.

\printbibliography

\end{document}